\documentclass[letterpaper]{article}

\PassOptionsToPackage{table}{xcolor}
\usepackage{template,times}
\usepackage{hyperref}
\usepackage{url}
\usepackage{graphicx}
\usepackage{algorithm}
\usepackage{algorithmic}
\usepackage{booktabs}
\usepackage{rotating}
\usepackage{enumitem}
\usepackage{orcidlink}
\usepackage{subcaption}
\usepackage{amsmath, amssymb}
\usepackage{lineno}
\usepackage{pifont}
\usepackage{xcolor}
\setcitestyle{numbers,square,citesep={,}}

\finalcopy

\title{SemiOccam: A Robust Semi-Supervised Image Recognition Network Using Sparse Labels}

\author{\hspace*{-0.8mm}
Rui Yann\thanks{Rui performed this research during an internship.}\ \ \orcidlink{0009-0004-0474-4522} \\
\href{mailto:Shu1L0n9@gmail.com}{\texttt{Shu1L0n9@gmail.com}}
\And
Tianshuo Zhang\ \ \orcidlink{0000-0002-7920-4473} \\
\href{mailto:zhang.tianshuo@163.com}{\texttt{zhang.tianshuo@163.com}}
\And
Xianglei Xing\thanks{Corresponding author.}\ \ \orcidlink{0000-0002-4159-1922} \\
\href{mailto:xingxl@hrbeu.edu.cn}{\texttt{xingxl@hrbeu.edu.cn}}
\AND
General Artificial Intelligence Laboratory \\
College of Intelligent Systems Science and Engineering \\
Harbin Engineering University \\
Harbin, 150001, China \\
}

\begin{document}

\maketitle

\begin{figure}[ht]
    \centering
    \includegraphics[width=0.95\textwidth]{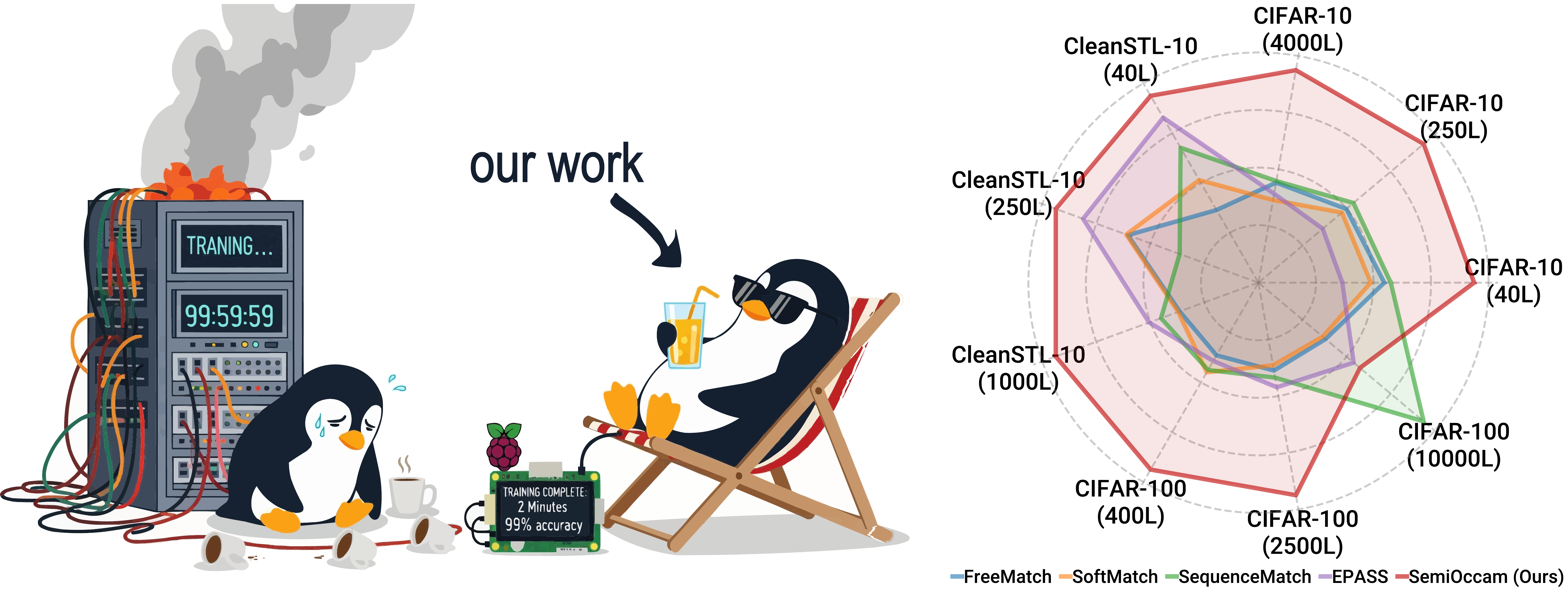}
    \label{fig:ssl_performance_stl10}
\end{figure}

\begin{abstract}
We present \textbf{SemiOccam}, an image recognition network that leverages semi-supervised learning in a highly efficient manner. \textit{Existing works} often rely on complex training techniques and architectures, \textit{requiring hundreds of GPU hours for training}, while their \textit{generalization ability} when dealing with extremely limited labeled data \textit{remains to be improved}. To address these limitations, we construct a hierarchical mixture density classification decision mechanism by \textbf{optimizing mutual information} between feature representations and target classes, compressing redundant information while retaining crucial discriminative components. Experimental results demonstrate that our method achieves \textbf{state-of-the-art performance} on \textbf{three commonly used datasets}, with accuracy \textbf{exceeding 95\%} on two of them, using \textbf{only 4 labeled samples per class}, and its simple architecture \textbf{keeps training time to minute-level}. Notably, this paper \textbf{reveals a long-overlooked data leakage issue} in the STL-10 dataset for semi-supervised learning tasks and removes duplicates to ensure the reliability of experimental results. We also release the deduplicated CleanSTL-10 dataset to \textbf{facilitate fair and reliable research} in future semi-supervised learning.
\ifpaperfinal\footnote{Code available at \url{https://github.com/Shu1L0n9/SemiOccam}.}\fi
\ifpaperfinal\footnote{CleanSTL-10 available at \url{https://huggingface.co/datasets/Shu1L0n9/CleanSTL-10}.}\fi
\end{abstract}


\section{Introduction}\label{sec:introduction}
Deep learning has achieved remarkable success in image classification tasks. However, its performance is \textbf{highly dependent} on large-scale annotated datasets. In real-world applications, labeled data is often scarce and \textbf{expensive to acquire}. Therefore, semi-supervised learning (SSL) has received widespread attention in recent years as an effective solution. The goal of SSL is to train models using a small amount of labeled data and large amount of unlabeled data, thereby improving the model's generalization ability. Nevertheless, most existing SSL methods, such as consistency regularization and pseudo-labeling, rely on \textbf{complex training strategies and network designs}, leading to \textbf{training costs of hundreds of GPU hours}~\cite{sohn2020fixmatch}\cite{zhang_flexmatch_2021}\cite{wang2022usb} even on high-performance computing resources like NVIDIA V100, and their \textbf{generalization ability when labeled data are extremely limited still needs improvement}. Moreover, although some recent works have explored Vision Transformers (ViT) in SSL~\cite{zhao2023learning}, they have yet to fully exploit its advantages.

The inspiration for this research stems from the \textbf{principle of Occam's razor}. This principle posits that among competing hypotheses that explain the data well, the simplest one is preferred. This motivates us to design a solution that balances model complexity and interpretability, ensuring efficient and effective feature representation while preventing unnecessary overfitting.

Aligned with this ethos, we delve into Gaussian Mixture Models (GMM), powerful probabilistic models that assume data is generated from a mixture of multiple Gaussian distributions. This exploration also draws from the systematic study by \citeauthor{miller_mixture_1996} on the "Mixture of Experts (MoE)" structure based on the joint probability model of features and labels. In our research, we found that the performance of GMM is highly related to the quality of feature vectors. If the features lack representativeness or have insufficient discriminative power, the classification performance will be severely affected.

\begin{figure}[ht]
    \centering
    \includegraphics[width=0.85\textwidth, trim=12mm 12mm 6mm 10mm, clip]{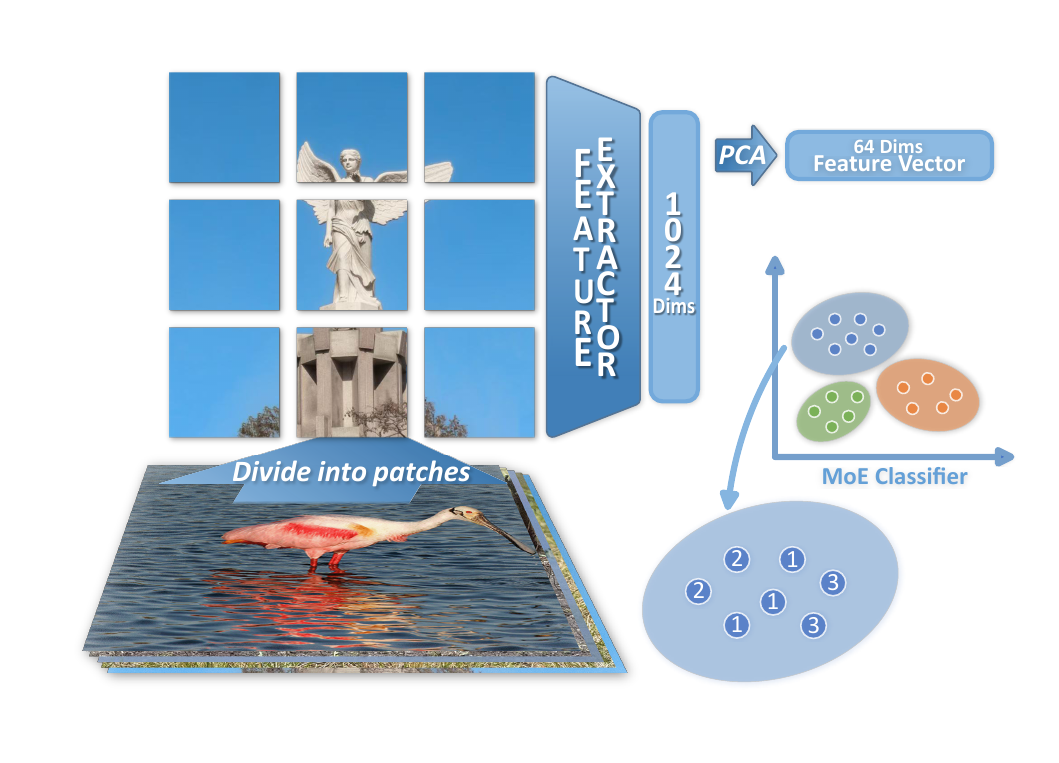}
    \caption{Overview of our SemiOccam network. The network consists of a feature extractor and a semi-supervised mixture of experts classifier. An example of a sub-model is shown in the ellipse at the bottom right, where the circle labeled \ding{172} represents a feature vector with a corresponding label, \ding{173} represents a pseudo-labeled feature vector, and \ding{174} represents an unlabeled feature vector.}
    \label{fig:overview}
\end{figure}

\textbf{To address this}, we designed a feature encoding framework based on a self-attention mechanism to fully exploit the global contextual information in images and generate more discriminative feature representations. This enhanced feature quality is crucial for optimizing MoE classifier's performance. \textbf{Furthermore}, to effectively leverage large amounts of unlabeled data, we incorporated a semi-supervised learning strategy utilizing a pseudo-labeling mechanism. An overview of our network, illustrating the integration of these approaches, is provided in Figure \ref{fig:overview}.

This approach achieves superior classification performance within the framework of mixture density estimation, significantly \textbf{overcomes the limitations of traditional methods} in terms of \textit{training cost} and \textit{model generalization performance}. Taking the Dino with ViT-large backbone as our feature extraction model, we validated the proposed semi-supervised gaussian mixture model (SGMM) method on various datasets, results show that our method \textbf{achieves state-of-the-art performance within minutes on a single Tesla T4 accelerator} and exhibits good stability and robustness, enabling \textbf{fast iteration} and potential for \textbf{edge deployment}.

It is worth mentioning that we have revealed a serious problem that has been overlooked by existing semi-supervised learning research: there are \textbf{7,500+ duplicate samples between the training samples and the testing samples in STL-10}, which is undoubtedly a serious data leakage problem when directly used for semi-supervised learning tasks. To address this, we \textbf{release the deduplicated CleanSTL-10 dataset} for fair and reliable evaluation.

Our \textbf{main contributions} can be summarized as follows:

\begin{itemize}[leftmargin=4pt, itemsep=4pt]
    \item[] We \textbf{\textcolor[HTML]{6E9B68}{introduce}} the powerful ViT as a feature extractor into the classical yet often underappreciated MoE classifier for image recognition tasks.
    \item[] We \textbf{\textcolor[HTML]{8F7CC5}{propose}} a unified framework that effectively exploits unlabeled data by seamlessly integrating deep representation extraction and semi-supervised learning.
    \item[] We \textbf{\textcolor[HTML]{D87F7F}{reveal}} the longstanding, yet overlooked issue of training set and test set contamination in the STL-10 dataset, which has not been addressed in other works.
    \item[] We \textbf{\textcolor[HTML]{E6B06B}{release}} the deduplicated CleanSTL-10 dataset to promote fair and reliable future research.
\end{itemize}

The \textbf{remainder of this paper} is organized as follows: Section \ref{sec:related-works} reviews related works; Section \ref{sec:method} introduces our SemiOccam network; Section \ref{sec:experiments} demonstrates and analyzes experimental results, including both classification performance and training cost; Section \ref{sec:conclusions} discusses the significance and limitations of the research findings and outlines future research directions.

\section{Related Works}\label{sec:related-works}

\subsection{Semi-Supervised Learning and Gaussian Mixture Models}\label{subsection:semi-supervised-learning-and-gmm}

\textbf{Semi-Supervised Learning} has been widely applied in tasks such as image classification and speech recognition in recent years, especially in scenarios where labeled data is scarce. Early SSL methods mainly relied on generating pseudo-labels and consistency regularization.

\textbf{Pseudo-Labeling} is one of the classic methods in SSL, with the core idea of using an existing model to predict unlabeled data \cite{yarowsky_unsupervised_1995} and adding its predicted labels to the training set as pseudo-labels. Representative works include \cite{lee_pseudo-label_2013} and \cite{zhang_flexmatch_2021}, which iteratively enhance the labeled dataset through self-training. Although pseudo-labeling methods can effectively improve model performance, they are often affected by the quality of pseudo-labels and can easily lead to model overfitting. The study by \cite{mishra2024trusttrustmiscalibrationsemisupervised} explores the issue of overconfidence caused by pseudo-labels in semi-supervised learning, analyzes the source of calibration errors, and aims to improve the reliability of the model by evaluating and potentially improving methods.

\textbf{Consistency Regularization} methods improve the generalization ability of models by forcing them to maintain consistency on unlabeled data. The Virtual Adversarial Training method proposed by \cite{miyato2018virtual} enhances the robustness of the model by perturbing the input data and maintaining the stability of the prediction results. The Mean Teacher method proposed by \cite{tarvainen2017mean} also further improves model performance by using the consistency between the predictions of the teacher model and the output of the student model. Although these methods have achieved good results in many tasks, they often rely on complex training strategies and architectural designs, and their generalization ability across different datasets still needs to be improved.

\textbf{Semi-Supervised Gaussian Mixture Model} is the basic probabilistic framework of our method. It combines GMM with unlabeled data and proposes a learning method that simultaneously considers both labeled and unlabeled data by maximizing the joint log-likelihood function. Our method is inspired by the classifier structure and learning algorithm proposed by \citeauthor{miller_mixture_1996}, which effectively leverages unlabeled data to improve performance. In addition, many impressive works, such as \cite{zong_deep_2018}, have extended GMM by incorporating deep autoencoders for tasks like anomaly detection.

Most related to ours is the model of \cite{zhao2023learning}, which also uses ViT for image embedding and then applies GMM for classification. This model is very similar to SemiOccam, but it addresses the generalized category discovery problem from \cite{vaze2022generalized}, aiming to discover new categories, whereas our work goes further to demonstrate that combining deep learning with GMM can outperform state-of-the-art methods, especially in scenarios with extremely limited labeled data, yielding significant advantages.

\subsection{Deep neural networks and Visual Transformers}\label{subsection:dnn-and-vit}

\textbf{Deep Neural Networks} have made significant progress in various supervised learning tasks in recent years. The breakthrough achieved by AlexNet, proposed by \cite{krizhevsky_imagenet_2012}, in image classification tasks laid the foundation for the development of deep learning. To improve the generalization ability of Deep Neural Networks (DNN), many methods have been proposed to address overfitting issues, such as Dropout by \cite{srivastava_dropout_2014} and Batch Normalization by \cite{ioffe_batch_2015}. The Residual Network (ResNet) proposed by \cite{he2016deep} effectively solved the gradient vanishing problem in deep network training by introducing residual connections, making it possible to train deeper networks. Subsequently, DLA by \cite{yu2018deep} introduced a dynamic hierarchical attention mechanism on top of ResNet, further optimizing layer-wise interaction and feature fusion.

\textbf{Visual Transformers} are image classification models based on the Transformer architecture that capture global information in images through self-attention mechanisms. \cite{dosovitskiy_image_2021} proposed dividing images into fixed-size patches and applying self-attention mechanisms. In this way, ViT demonstrates strong performance in image classification tasks, especially on large-scale datasets, exhibiting higher generalization ability compared to traditional convolution neural network (CNN) architectures. The success of ViT has inspired many researchers to try applying it to self-supervised learning. The DINO method proposed by \cite{caron_emerging_2021} learns useful feature representations through self-supervised pre-training of ViT models. These representations can be fine-tuned on a small amount of labeled data to achieve high classification accuracy. 

Similar to our work, these two methods integrate ViT with machine learning techniques. \cite{wang2023p2fe} combines CNN and ViT to better capture both local and global features, while \cite{roy2025simpoolformer} enhances hyperspectral image classification by refining the attention mechanism and adding an auxiliary branch. Both approaches optimize the ViT architecture to improve performance and efficiency in specific tasks.

\subsection{Self-Supervised Learning and Data Leakage Issues}\label{subsection:self-supervised-learning-and-data-leakage}

\textbf{Self-Supervised Learning} is an unsupervised learning approach that automatically learns feature representations through pretext tasks (e.g., image jigsaw puzzles, predicting future frames). The CLIP model proposed by \cite{radford_learning_2021} achieves strong image understanding capabilities through self-supervised learning of image-text contrast, making its performance on unlabeled data very prominent.

In addition, methods like SimCLR from \cite{chen_simclr_2020} and MoCo from \cite{He_2020_CVPR} train neural networks through contrastive learning, enabling them to learn discriminative features without labeled data. These methods provide higher quality representations for feature extractors, thereby improving the performance of downstream tasks (such as classification, face recognition, etc.).

\textbf{Data leakage} issue has always been a potential challenge in machine learning. Off-topic images, near-duplicate images, and labeling errors in benchmark datasets can lead to inaccurate estimations of model performance. Handling and removing duplicate samples has become an issue that cannot be ignored in semi-supervised learning. Therefore, \citeauthor{grogerintrinsic}'s work re-examines the task of data cleaning and formalizes it as a ranking problem.

\section{Method}\label{sec:method}

We assume that the entire dataset $\mathcal{X} = \{\mathbf{x}_i \in \mathbb{R}^{H \times W \times C}, i = 1, \ldots, N\}$ consists of a labeled dataset $\mathcal{X}_l = \{(\mathbf{x}_1, c_1), \ldots, (\mathbf{x}_{N_l}, c_{N_l})\}$ and an unlabeled dataset $\mathcal{X}_u = \{\mathbf{x}_{N_l+1}, \ldots, \mathbf{x}_N\}$. Here, the number of labeled data $N_l$ is much smaller than the total data amount $N$ ($N_l \ll N$).

Our method is as follows:

\subsection{Powerful Feature Extractor}\label{subsection:feature_extractor}

For each input image $\mathbf{x}_i \in \mathcal{X}$, where $\mathbf{x}_i \in \mathbb{R}^{H \times W \times C}$, we divide it into a sequence of fixed-size $P \times P$ patches. Let $N_p = \frac{HW}{P^2}$ denote the number of patches per image. Each patch $\mathbf{p}^{(i)}_k \in \mathbb{R}^{P^2C}$ (where $k = 1, \ldots, N_p$) is flattened into a vector. Then, through a linear transformation, each patch vector $\mathbf{p}^{(i)}_k$ is mapped to a $D$-dimensional vector $\mathbf{z}^{(i)}_k \in \mathbb{R}^{D}$, expressed as:

\begin{figure}[H]
    \centering
    \includegraphics[width=1\textwidth]{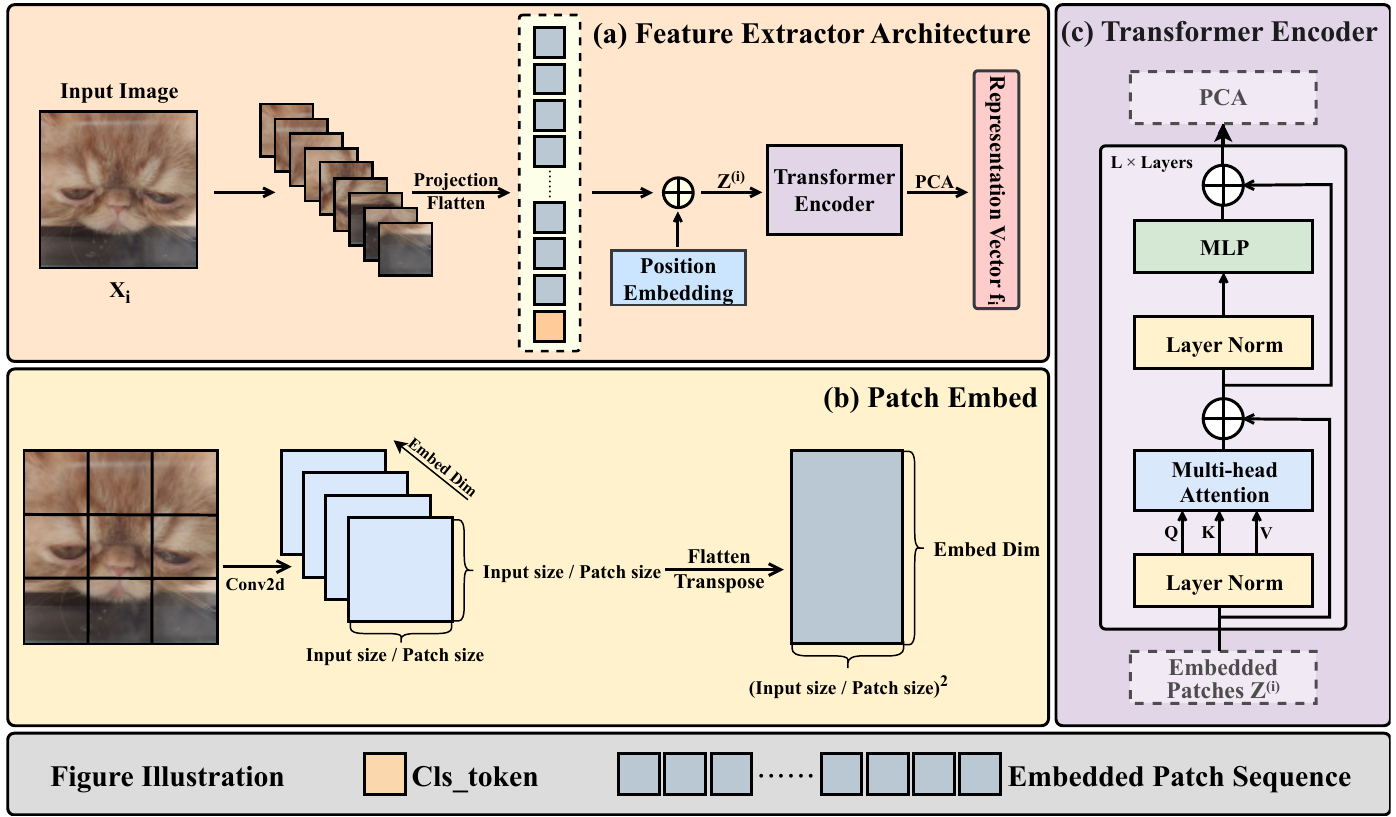}
    \caption{Feature Extraction Architecture. This diagram illustrates the ViT architecture specifically configured as a feature extractor. Input images are processed through patch embedding and a Transformer Encoder, yielding high-dimensional feature vectors for downstream tasks.}
    \label{fig:Feature extractor}
\end{figure}

\begin{equation}
\mathbf{z}^{(i)}_k = \mathbf{W}\mathbf{p}^{(i)}_k + \mathbf{b}
\label{eq:patch_embedding}
\end{equation}

where $\mathbf{W} \in \mathbb{R}^{D \times P^2C}$ and $\mathbf{b} \in \mathbb{R}^D$ are learnable parameters. A learnable positional embedding $\mathbf{e}_k \in \mathbb{R}^D$ is added to each patch vector $\mathbf{z}^{(i)}_k$ to retain the positional information of the image patches, yielding $\mathbf{z}'^{(i)}_k = \mathbf{z}^{(i)}_k + \mathbf{e}_k$. This results in the image patch sequence $\mathbf{Z}^{(i)} = \{ \mathbf{z}'^{(i)}_1, \mathbf{z}'^{(i)}_2, \ldots, \mathbf{z}'^{(i)}_{N_p} \}$.

Subsequently, the image patch sequence $\mathbf{Z}^{(i)}$ is input into a Transformer encoder. The encoder utilizes the self-attention mechanism, which is defined as:

\begin{equation}
\text{Attention}(\mathbf{Q}, \mathbf{K}, \mathbf{V}) = \text{softmax}\left(\frac{\mathbf{Q}\mathbf{K}^T}{\sqrt{D}}\right)\mathbf{V}
\label{eq:self_attention}
\end{equation}

where $\mathbf{Q} = \mathbf{Z}^{(i)}\mathbf{W}_Q$, $\mathbf{K} = \mathbf{Z}^{(i)}\mathbf{W}_K$, and $\mathbf{V} = \mathbf{Z}^{(i)}\mathbf{W}_V$. Through multiple layers of Transformer encoding and Principal Component Analysis (PCA) dimensionality reduction, we finally obtain high-quality image feature vectors $\mathbf{f}_i \in \mathbb{R}^d$ where $d < D$. The overall feature extraction pipeline is illustrated in Figure~\ref{fig:Feature extractor}.

\subsection{Semi-supervised Mixture of Experts Classifier}\label{subsection:sgmm_classifier}

Each feature vector $\mathbf{f}_i \in \mathbb{R}^d$ is obtained from the feature extractor. Each labeled sample $(\mathbf{f}_i, c_i)$ is associated with a class label $c_i$ belonging to the label set $\mathcal{L} = \{1, 2, \ldots, K\}$, where $K = \max\{c_1, \ldots, c_{N_l}\}$ represents the total number of classes.

\begin{figure}[H]
    \centering
    \includegraphics[width=0.85\textwidth, trim=5mm 0mm 0mm 3mm, clip]{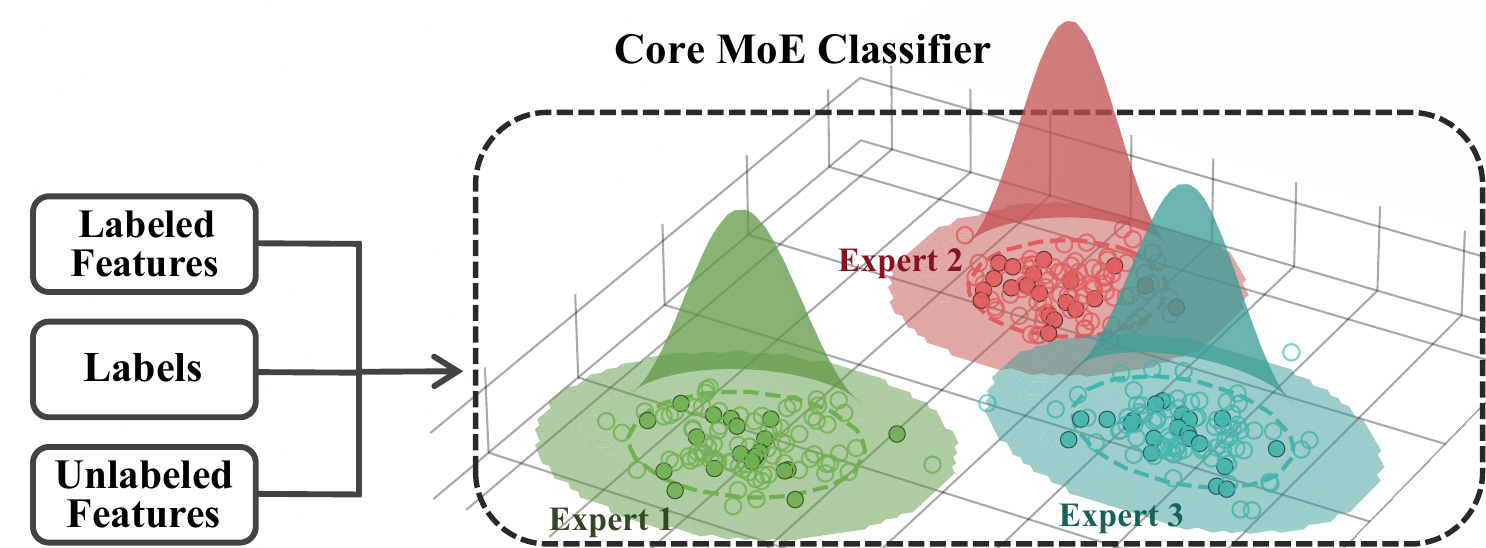}
    \caption{Illustration of our Classifier. The figure visualizes a model with three gaussian components. The solid spheres represent labeled features, while the hollow spheres denote unlabeled features.}
    \label{fig:moe_classifier}
\end{figure}

To exploit both labeled and unlabeled data, we formulate a semi-supervised maximum likelihood estimation objective. The overall objective is:

\begin{equation}
\mathcal{L}(\Theta) = \mathcal{L}_1(\Theta) + \lambda \mathcal{L}_2(\Theta)
\end{equation}

where $\mathcal{L}_1$ corresponds to the supervised loss, $\mathcal{L}_2$ to the unsupervised loss, and $\lambda$ balances the two terms. In our implementation, we set $\lambda = 1$ for simplicity.

Specifically,

\begin{equation}
\mathcal{L}_1(\Theta) = \sum_{\mathbf{f}_i \in \mathcal{F}_l} \log p_{\Theta}(\mathbf{f}_i, c_i), \qquad \mathcal{L}_2(\Theta) = \sum_{\mathbf{f}_i \in \mathcal{F}_u} \log p_{\Theta}(\mathbf{f}_i)
\end{equation}

The joint likelihood of all samples can thus be written as:

\begin{equation}
\mathcal{L}(\Theta) = \underbrace{\sum_{\mathbf{f}_i \in \mathcal{F}_l} \log \sum_{l=1}^L \pi_l p(\mathbf{f}_i \vert \theta_l) P(c_i \vert l)}_{\text{labeled data likelihood}} + \underbrace{\sum_{\mathbf{f}_i \in \mathcal{F}_u} \log \sum_{l=1}^L \pi_l p(\mathbf{f}_i \vert \theta_l)}_{\text{unlabeled data likelihood}}
\label{eq:sgmm}
\end{equation}

where $\Theta = \{\pi_l, \boldsymbol{\mu}_l, \boldsymbol{\Sigma}_l, P(c \vert l) \}_{l=1 \dots L, c=1 \dots K}$ is the set of model parameters with $\sum_{l=1}^L \pi_l = 1$, $\pi_l$ represents the mixing coefficient of the $l$-th gaussian component, $p(\mathbf{f}_i \vert \theta_l)$ is the probability density function of the $l$-th gaussian component with parameters $\theta_l = \{\boldsymbol{\mu}_l, \boldsymbol{\Sigma}_l\}$, and $P(c_i \vert l)$ represents the conditional probability that a feature belongs to class $c_i$ given gaussian component $l$.

The EM algorithm updates parameters as follows:

\begin{equation}
\gamma_{il}^{(t)} = \frac{\pi_l^{(t)} p(\mathbf{f}_i \vert \theta_l^{(t)})}{\sum_{j=1}^L \pi_j^{(t)} p(\mathbf{f}_i \vert \theta_j^{(t)})} \quad \forall \mathbf{f}_i \in \mathcal{F}_u
\label{eq:gamma_il}
\end{equation}

\begin{equation}
\gamma_{il \vert c_i}^{(t)} = \frac{\pi_l^{(t)} p(\mathbf{f}_i \vert \theta_l^{(t)}) P(c_i \vert l)^{(t)}}{\sum_{j=1}^L \pi_j^{(t)} p(\mathbf{f}_i \vert \theta_j^{(t)}) P(c_i \vert j)^{(t)}} \quad \forall \mathbf{f}_i \in \mathcal{F}_l
\label{eq:gamma_il_ci}
\end{equation}

\begin{equation}
\boldsymbol{\mu}_l^{(t+1)} = \frac{\sum_{\mathbf{f}_i \in \mathcal{F}_l} \mathbf{f}_i \gamma_{il \vert c_i}^{(t)} + \sum_{\mathbf{f}_i \in \mathcal{F}_u} \mathbf{f}_i \gamma_{il}^{(t)}}{\sum_{\mathbf{f}_i \in \mathcal{F}_l} \gamma_{il \vert c_i}^{(t)} + \sum_{\mathbf{f}_i \in \mathcal{F}_u} \gamma_{il}^{(t)}}
\label{eq:mu_l}
\end{equation}

\begin{equation}
\boldsymbol{\Sigma}_l^{(t+1)} = \frac{\sum_{\mathbf{f}_i \in \mathcal{F}_l} M_{il}^{(t)} \gamma_{il \vert c_i}^{(t)} + \sum_{\mathbf{f}_i \in \mathcal{F}_u} M_{il}^{(t)} \gamma_{il}^{(t)}}{\sum_{\mathbf{f}_i \in \mathcal{F}_l} \gamma_{il \vert c_i}^{(t)} + \sum_{\mathbf{f}_i \in \mathcal{F}_u} \gamma_{il}^{(t)}}
\label{eq:sigma_l}
\end{equation}
where $M_{il}^{(t)} = (\mathbf{f}_i - \boldsymbol{\mu}_l^{(t)})(\mathbf{f}_i - \boldsymbol{\mu}_l^{(t)})^\top$.

\begin{equation}
\pi_l^{(t+1)} = \frac{1}{N} \left( \sum_{\mathbf{f}_i \in \mathcal{F}_l} \gamma_{il \vert c_i}^{(t)} + \sum_{\mathbf{f}_i \in \mathcal{F}_u} \gamma_{il}^{(t)} \right)
\label{eq:pi_l}
\end{equation}

\begin{equation}
P(k \vert l)^{(t+1)} = \frac{\sum_{\mathbf{f}_i \in \mathcal{F}_l, c_i=k} \gamma_{il \vert c_i}^{(t)}}{\sum_{\mathbf{f}_i \in \mathcal{F}_l} \gamma_{il \vert c_i}^{(t)}}
\label{eq:P_k_given_l}
\end{equation}

The model learns its optimal parameters, $\Theta$, by iterating between the Expectation and Maximization until the algorithm converges. Figure~\ref{fig:moe_classifier} offers a visual illustration of the final learned state, presenting a classifier made up of three gaussian components.

\subsection{Pseudo-labeling Mechanism}\label{subsection:pseudo_labeling}

To effectively leverage information from the unlabeled dataset $\mathcal{X}_u$, we introduce a pseudo-labeling mechanism based on the \textit{component-class association probabilities}. Unlike traditional iterative methods, our pseudo-label generation is performed in a single calculation after the initial SGMM training reaches convergence, reducing computational cost and mitigating error accumulation risks.

For unlabeled features $\mathbf{f}_i \in \mathcal{F}_u$, we compute class probabilities through gaussian component responsibilities:

\begin{equation}
p_i^{(k)} = \sum_{l=1}^L P(k|l)^{(t)}\gamma_{il}^{(t)},\quad \xi_i = \max_{1 \leq k \leq K} p_i^{(k)}
\end{equation}

where $P(k|l)$ represents the class probability conditioned on component $l$, and $\gamma_{il}^{(t)}$ is the responsibility from Equation~\ref{eq:gamma_il}.

We construct class-specific candidate sets with confidence thresholding sorted by $\xi_i$ descending:

\begin{equation}
\mathcal{C}_k = \left\{ \mathbf{f}_i \mid \arg\max_k p_i^{(k)} = k,\ \xi_i > \tau, \quad \tau \in (0,1) \right\}
\label{eq:constuct_set_function}
\end{equation}

To prevent class imbalance, we adopt proportional sampling:

\begin{equation}
n^{\text{final}} = \min_{1 \leq k \leq K} \lfloor \alpha |\mathcal{C}_k| \rfloor \quad (\alpha \in (0,1))
\label{eq:constuct_set_function2}
\end{equation}

The pseudo-labeled set is then constructed as:

\begin{equation}
\mathcal{D}_p = \{ (\mathbf{f}_i, k) \mid k=1,\ldots,K,\ \mathbf{f}_i \in \mathcal{C}_k[1:n^{\text{final}}] \}
\label{eq:constuct_set_function3}
\end{equation}

Finally, combine the pseudo-labeled data \(\mathcal{D}_p\) with the labeled features \(\mathcal{F}_l\) and use the EM algorithm for iterative. We use Kmeans++ to initialize the parameters first, during the training process, log-likelihood value calculated using Equation \ref{eq:sgmm} increases steadily. We show the procedure of our method in Algorithm \ref{alg:vit_sgmm}.

\begin{algorithm}[H]
    \caption{SemiOccam Network}
    \label{alg:vit_sgmm}
    \begin{algorithmic}[1]
        \REQUIRE Labeled data $\mathcal{X}_l$, Unlabeled data $\mathcal{X}_u$, Pre-trained ViT, SGMM components $L$, Confidence threshold $\tau$, Sampling ratio $\alpha$
        \ENSURE Trained SGMM parameters $\Theta$
        
        \STATE Extract features $\mathcal{F}_l \leftarrow \{\text{ViT}(\mathbf{x}_i) | (\mathbf{x}_i, c_i) \in \mathcal{X}_l\}$ and $\mathcal{F}_u \leftarrow \{\text{ViT}(\mathbf{x}_i) | \mathbf{x}_i \in \mathcal{X}_u\}$
        \STATE PCA $\mathcal{F}_l$ and $\mathcal{F}_u$ into $\mathbb{R}^d$
        \STATE $L$ centers cluster $\mathcal{F}_l \cup \mathcal{F}_u$ with K-means++
        \STATE Initialize $\{\pi_l, \boldsymbol{\mu}_l, \boldsymbol{\Sigma_l}\}$ from cluster results
        
        \FOR{$t = 1$ to $T_1$}
            \STATE EM algorithm: Initial iterative via (\ref{eq:gamma_il}-\ref{eq:P_k_given_l})
        \ENDFOR
        
        \STATE Construct $\mathcal{D}_p$ using (\ref{eq:constuct_set_function}-\ref{eq:constuct_set_function3}) with $\tau$ and $\alpha$
        \STATE Augment labeled set: $\mathcal{F}_l \leftarrow \mathcal{F}_l \cup \mathcal{D}_p$
        
        \FOR{$t = 1$ to $T_2$}
            \STATE EM algorithm: Final iterative via (\ref{eq:gamma_il}-\ref{eq:P_k_given_l})
        \ENDFOR
        
        \RETURN $\Theta$
    \end{algorithmic}
\end{algorithm}

\section{Experiments}\label{sec:experiments}

\subsection{Standard SGMM Performance Analysis}\label{subsection:standard_sgmm}

The performance of the SGMM is influenced by several factors. This section will delve into the impact of the number of labels, feature dimensions, and the number of gaussian components on the model's behavior.

SGMM's core of its labeling efficiency lies in using unlabeled data to assist the model in learning the data distribution. Theoretically, as the number of labeled samples increases, the model can more accurately estimate the conditional probability of categories \(P(k \vert l)\), thereby improving discrimination ability. When the labeled samples reach a certain scale, the marginal benefit of the additional information provided by unlabeled data diminishes, and the performance improvement curve tends to flatten. Therefore, SGMM's advantage is that it can quickly improve performance with very few labels and demonstrate high efficiency in applications with limited labeling resources.

\begin{figure}[htbp]
    \centering
    \includegraphics[width=\textwidth]{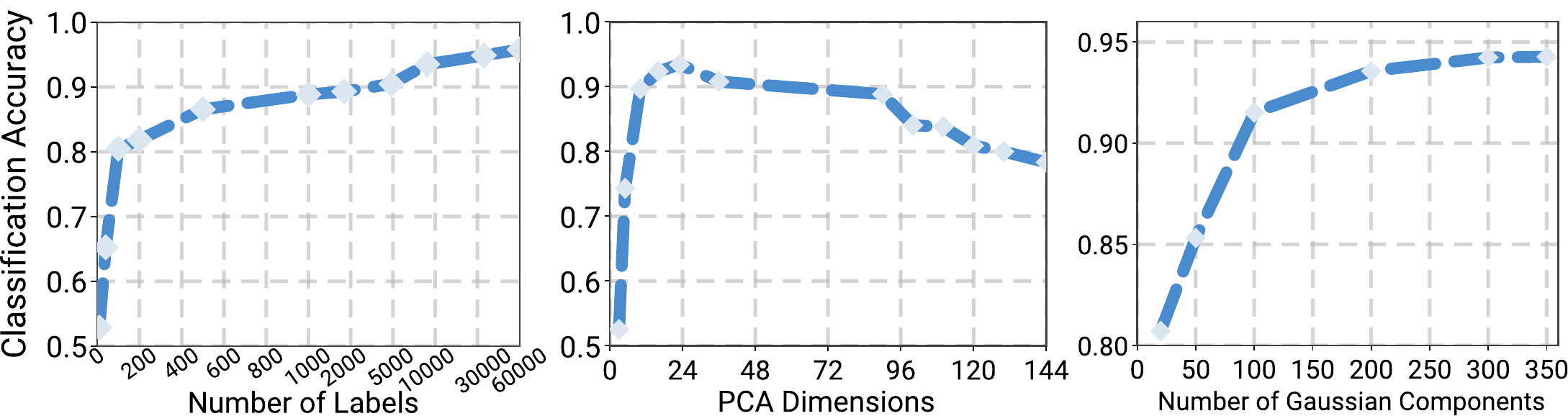}
    \caption{Relationship between number of labels/PCA dimensions/number of gaussian components and performance on the MNIST dataset. The first figure shows a steady improvement in accuracy with an increase in labels. The second figure shows the trend of SGMM accuracy with PCA dimensions, while the right figure shows the trend of accuracy with the number of gaussian submodels.}
\end{figure}

When dealing with high-dimensional data, the feature dimension is crucial to the performance and computational efficiency of SGMM. High-dimensional feature spaces can lead to the ``curse of dimensionality'', increasing the complexity of model training and the risk of overfitting, while also consuming excessive computational resources. Reducing feature dimensions through dimensionality reduction techniques such as PCA can retain discriminative feature information while removing redundant dimensions, alleviating these problems to some extent. However, excessive dimensionality reduction can lead to the loss of key feature information. Therefore, the feature dimension should be reduced as much as possible while ensuring model performance.

Gaussian Mixture Models use multiple gaussian components to fit complex data distributions, and the number of components directly affects the model's complexity and fitting ability. When the number of components is small, the model may not be able to fully capture the fine-grained structure of the data distribution, leading to underfitting. When the number of components is too large, the model complexity increases, which can easily lead to overfitting of the training data, reducing generalization ability and increasing computational costs. Therefore, the optimal number of gaussian components should match the intrinsic class structure and complexity of the dataset.

In summary, labeling efficiency reflects the core advantage of semi-supervised learning. Feature dimension efficiency emphasizes the role of dimensionality reduction in improving model efficiency and generalization ability. The number of gaussian components reflects the trade-off between model complexity and data fitting ability. Understanding the impact of these factors on SGMM performance helps to adjust model parameters according to specific tasks in practical applications.

\subsection{More Powerful Feature Extractors}\label{subsection:more_powerful_feature_extractors}

\begin{figure}[ht]
    \centering
    \includegraphics[width=\textwidth]{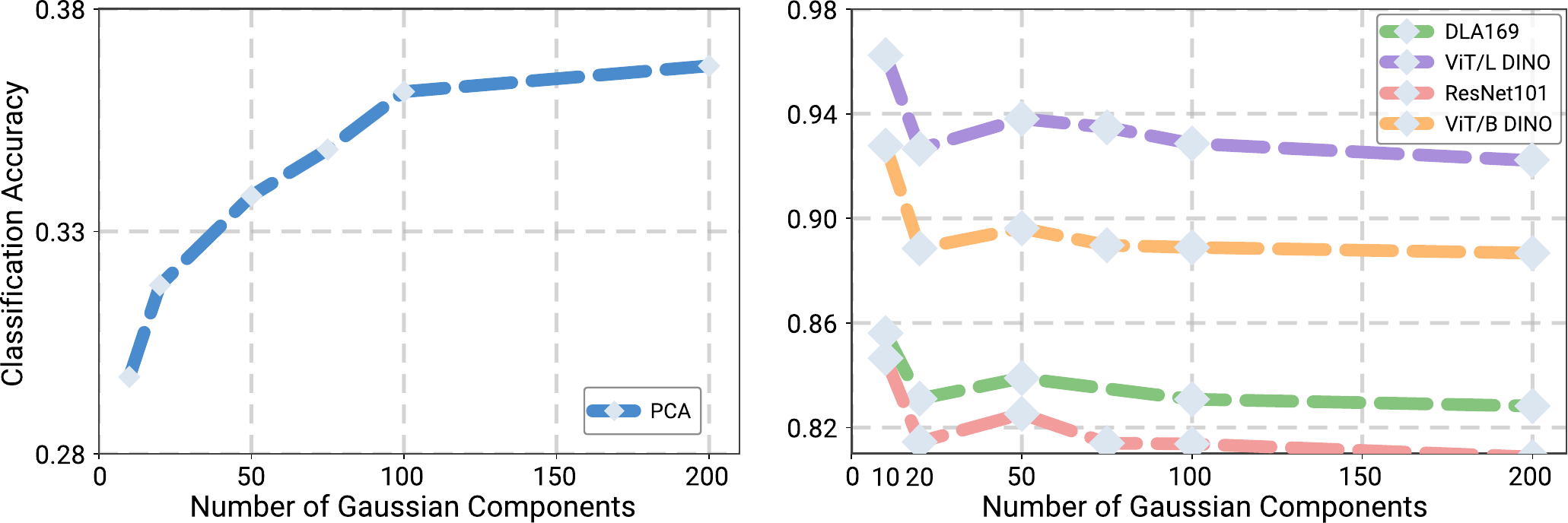}
    \caption{Performance trends. The left figure shows the accuracy trend of SGMM on the CIFAR-10 dataset processed with PCA, while the right figure shows the accuracy trend with the number of gaussian submodels when using DNNs and ViTs.}
    \label{fig:n_components_accuracy_cifar10_trend}
\end{figure}

To further analyze the performance of the SGMM, we experimented on the CIFAR-10 dataset using different feature extractors: PCA, DNNs (ResNet101, DLA169), and ViT-Base/Large backbone DINO pre-trained models. Figure \ref{fig:n_components_accuracy_cifar10_trend} shows that with DNNs and ViTs, increasing the number of gaussian components leads to a decrease in classification accuracy, which contrasts with the performance trend observed when using PCA.

\begin{figure}[ht]
    \centering
    \begin{minipage}[b]{0.32\textwidth}
        \centering
        \includegraphics[height=1\textwidth, trim=13.8mm 8.9mm 2.3mm 2.5mm, clip]{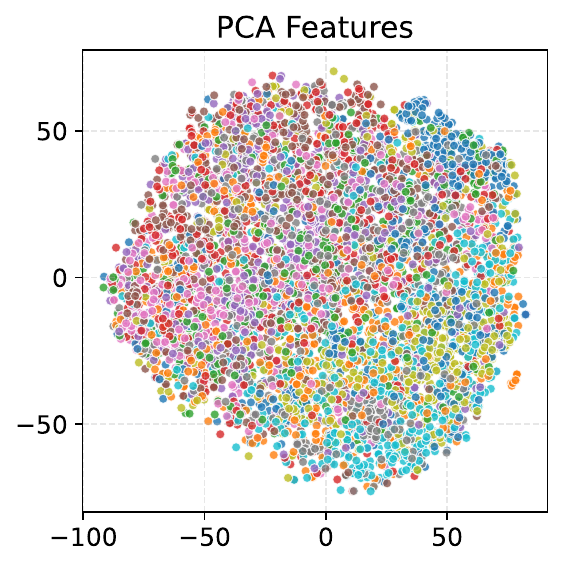}
    \end{minipage}
    \hfill
    \begin{minipage}[b]{0.32\textwidth}
        \centering
        \includegraphics[height=1\textwidth, trim=13.8mm 8.9mm 2.3mm 2.5mm, clip]{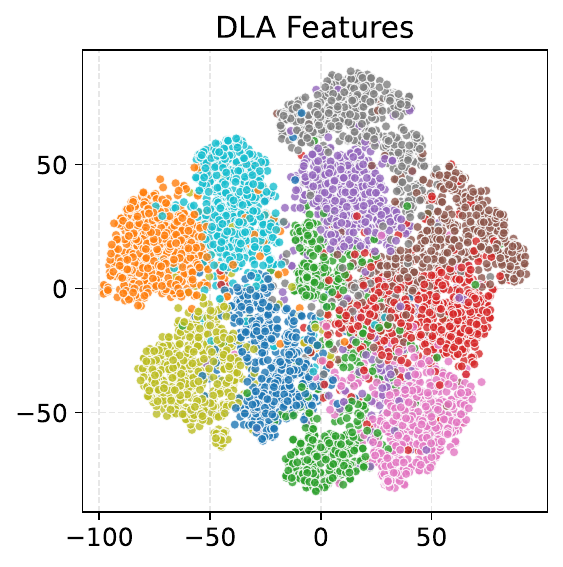}
    \end{minipage}
    \hfill
    \begin{minipage}[b]{0.32\textwidth}
        \centering
        \includegraphics[height=1\textwidth, trim=16.5mm 8.9mm 2.3mm 2.5mm, clip]{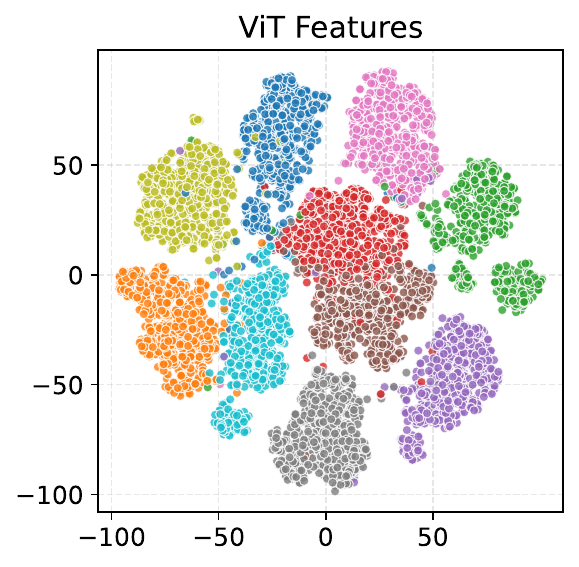}
    \end{minipage}
    \caption{t-SNE Visualization. t-SNE visualization results of different feature extraction methods on the CIFAR-10 dataset.}
    \label{fig:tsne_visualization}
\end{figure}

Next, we performed t-SNE visualization on the features extracted by the three methods. In Figure \ref{fig:tsne_visualization}, we present the t-SNE visualization results of feature representations obtained using different feature extraction methods on the CIFAR-10 dataset. It can be clearly seen from the figure that the feature points extracted by PCA exhibit a highly mixed state in the t-SNE space, and it is difficult to distinguish between samples from different categories. This indicates that the feature representation after PCA dimensionality reduction lacks sufficient discriminative power. In contrast, the feature points extracted using deep neural networks exhibit a certain clustering trend in the t-SNE space, but there is still some overlap between the boundaries of different categories. The feature points extracted using Visual Transformer form a relatively clear cluster structure in the t-SNE space, which indicates that the feature representation extracted by ViT has stronger separability and can better capture the intrinsic structure of the data, providing a more favorable feature basis for subsequent classification tasks.

PCA has limited feature discrimination ability, so it needs more gaussian sub-models to model complex intra-class variations. At this time, increasing the number of gaussian components helps to approximate the true distribution; while the features extracted by DNN/ViT are close to a single gaussian for each category, increasing the number of components will cause the model complexity to exceed the actual needs. These analyses are consistent with the experimental results in Figure \ref{fig:n_components_accuracy_cifar10_trend}.

\begin{figure}[ht]
    \centering
    \begin{minipage}[b]{1\textwidth}
        \centering
        \includegraphics[width=\textwidth]{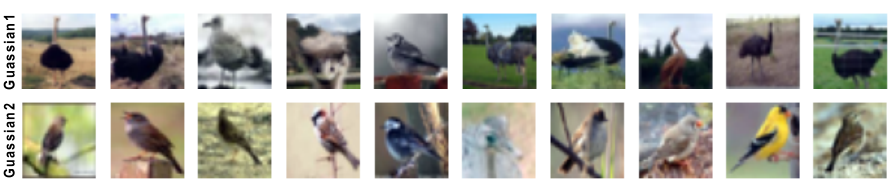}
    \end{minipage}
    \caption{Comparison raw images. Visualization of high-confidence samples from two gaussian components of the bird category on the CIFAR-10 dataset.}
    \label{fig:multiple_gaussian_visualization}
\end{figure}

We analyzed high-confidence samples from two gaussian components of the same bird category. Observing the original images in Figure \ref{fig:multiple_gaussian_visualization}, we found that these two gaussian components exhibit different feature preferences: one component tends to give high confidence to samples of ostriches (long-necked birds), while the other component prefers plump, neckless birds.

\begin{figure}[ht]
    \centering
    \begin{minipage}[b]{0.32\textwidth}
        \centering
        \includegraphics[width=\linewidth]{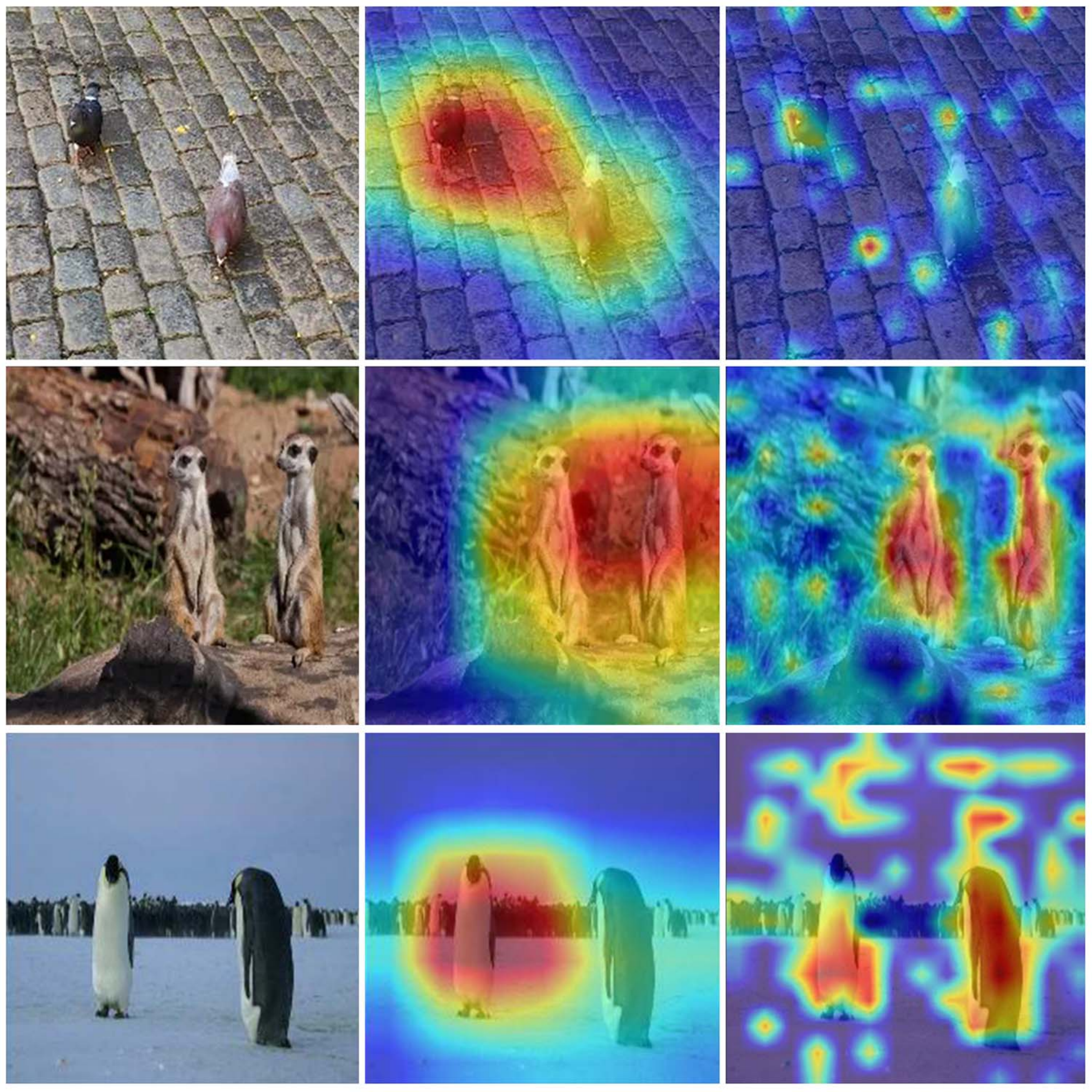}
    \end{minipage}%
    \hfill
    \begin{minipage}[b]{0.66\textwidth}
        \centering
        \includegraphics[width=\linewidth, trim=2.5mm 3mm 2.3mm 2mm, clip]{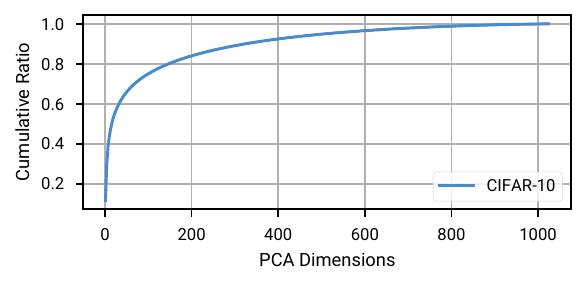}
    \end{minipage}
    \caption{Visual comparison of feature attention (left) and PCA variance analysis (right).}
    \label{fig:combined_heatmap_pca}
\end{figure}

We compare the heatmaps of DNN and ViT on several randomly selected images from the internet, each containing two animals simultaneously. The heatmaps are generated using the Grad-CAM method, which visualizes the importance of each pixel in the image for the classification result. As shown in Figure \ref{fig:combined_heatmap_pca}, the heatmaps of DNN and ViT differ significantly. While DNN focuses primarily on one animal, ViT attends to both animals in the image, demonstrating its ability to capture global context more effectively. This comparison highlights the differences in how these models prioritize regions of the image during classification.
Through experimental results and visualization analysis, we found that DNN and ViT can better capture the distribution characteristics of data compared to the traditional PCA dimensionality reduction method, thereby improving the classification performance of SGMM.

\subsection{STL-10 Data Clean}\label{sec:stl10_data_clean}

Duplicate samples in the STL-10 dataset can potentially bias model evaluation. To address this, we deduplicated the dataset. Specifically, we used a script to compute image hashes for STL-10 dataset, then removed 7,545 samples from the train set that were found to be duplicates of samples in the test set. \textbf{We release the deduplicated CleanSTL-10 dataset} to facilitate fair and reliable research. The pseudo-code for this deduplication process is provided in Algorithm \ref{alg:deduplication}.

\begin{algorithm}
    \caption{Deduplication of Dataset}
    \label{alg:deduplication}
    \begin{algorithmic}[2]
    \REQUIRE Main dataset, test dataset, batch size
    \ENSURE Deduplicated dataset, CSV file with duplicate information
  
    \STATE Build test set hash dictionary: Compute hash for each test image
    \STATE Initialize valid\_indices and duplicates as empty lists
  
    \FOR{each batch in main dataset}
        \FOR{each image in batch}
            \STATE Compute hash of the image
            \IF{hash exists in test set hash dictionary}
                \STATE Record duplicate sample details
            \ELSE
                \STATE Add current index to valid\_indices
            \ENDIF
        \ENDFOR
    \ENDFOR
  
    \IF{duplicates is not empty}
        \STATE Save duplicate details to CSV file
    \ENDIF
    \end{algorithmic}
\end{algorithm}

Following processing, the training set consists of 5,000 labeled samples and 90,455 unsupervised samples. This effectively addresses the problem of data leakage, thereby guaranteeing the credibility of the experimental findings.

\begin{table}[H]
    \centering
    \caption{Example pairs of training and test samples sharing the same hash value.}
    \label{tab:duplicate_indices}
    \resizebox{\textwidth}{!}{
    \begin{tabular}{l l l}
        \toprule
        \textbf{Train Index} & \textbf{Test Index} & \textbf{Hash} \\
        \midrule
        5008 & 2075 & e777922f14273159776a089a6f44d2bbb03f23a2b618f0914ef7ca442e770396 \\
        5025 & 4149 & d91a18f5a643aaf3569387641a6f40a92ec82a32a27a532625414dbaf48b4d25 \\
        5032 & 6188 & eadb8e0cd40606bb3e6c58f710abb4d9769e43b43724ef764e7a74a4d179fff4 \\
        5041 & 4228 & 5c1dee15a65d67508008125e0f1085382084a860e491c10f0f6c15d7c7d06a36 \\
        5045 & 6029 & f4197466cd0d97da2a2b079bc7458d7af8d1c89a7178d30d14668364037d165a \\
        5062 & 4980 & 82a4e1da419ea969f5bc07f9664e8974dee8c8161334d0ad9abc89f9545e8f41 \\
        5071 & 7166 & b445c22c7f8505e1ac09d6efa5b756daac42da5ae70acd8d672b9c164c4c8bbc \\
        5072 & 7091 & 9fb660ef9603189ca18787bcf60966bc8cb07f6ad121eeb3602cb543a5123830 \\
        5073 & 2803 & 7e3753d43ea8769a2818d947feea2beb84cf5e3cb090d5b31c820e6f6f4f0d8a \\
        5092 & 6820 & c360dd72cb9f9056302cafbc2f5ccfe9de66fddac0b8ed4d91ce93996dd98503 \\
        5108 & 5761 & 03a02e10956c65799a5f5d082d61cbaf1de414773c7c025da8bfbc5553c6fdf0 \\
        5121 & 2143 & 8e98699fcf6ec4e0b8765866248e7702c2f26f9fe2d1f469eeb0f51f54bb5afb \\
        5129 & 1287 & 363360e93d8997c1138d1e4a0fde65b7b08befd496b450df72989aac5e5b4edb \\
        5139 & 3879 & b2fa77caa967437f1a85c6071b2345d420da69780036d44f761591ad6d213ff6 \\
        5143 & 4701 & 29b8233f6d9a24cfbbdd83f23c838d59b1a113ada232c0991bd63e1fa97c533e \\
        5165 & 6844 & 46f4be3edd44d770964fb27f6777e77f5a93d1623921c9d630f9454908f9e2ac \\
        5169 & 4665 & ee864c70c6d01868bcde6eb61261c33057f39efbe9197a7590aed629a4215bae \\
        5172 &  461 & aa81e6c7a9a3c597b331abd352271f91eb4f1129b41cffd164b35371c41a8217 \\
        5188 & 2893 & 0da766b7fe175d6cb8328437019e4fa2d87f4878cdb09e9b2d001fe6374a100e \\
        5191 & 4131 & b909207b8b5822d2a2197ab4563fc32def9b9d04934285928a0ad6f08d1de9f4 \\
        5197 & 6931 & 013a41f75629ce371fbbec3f9a845aa5ec3e615819a4e4ae67b3dbe0d70ece0d \\
        5214 &  991 & 9a3201824495feabca12e1ef2339c410ff1f30c412ff35d85743bcd16f6bcabc \\
        5226 &  196 & 7309f100d4dc411520175d905e2cdebdd3ace6422ec0310efb61f7a93052402b \\
        \bottomrule
    \end{tabular}
    }
\end{table}

We present an example table of training and test sample indices with duplicate hash values in Table~\ref{tab:duplicate_indices}. The full list can be found in our project repository.

\subsection{Experiment Setup}\label{subsection:experiment_setup}

\paragraph{Datasets.}
We report results on the Cifar-10/100 and CleanSTL-10 datasets. CIFAR-10 contains 10 classes with 6,000 32x32 color images per class, totaling 60,000 images, of which 50,000 are for training and 10,000 for testing. CIFAR-100 is similar to CIFAR-10 but contains 100 classes with 600 images per class, of which 500 are for training and 100 for testing. STL-10 is an image classification dataset designed for self-supervised and semi-supervised learning, containing 10 classes with 5,000 labeled training samples and 100,000 unlabeled images, each at a resolution of 96×96 pixels. We created CleanSTL-10 by carefully removing duplicates and overlaps between training and test sets, as detailed in Section~\ref{sec:stl10_data_clean}. After this deduplication, CleanSTL-10 contains 90,455 training images and 8,000 testing images, ensuring more reliable and fair evaluation.

\paragraph{Implementation details.}
We use the DINO pre-trained model with a ViT-Large backbone to extract feature representations. For dimensionality reduction, PCA is applied with dimensions selected based on explained variance, typically preserving over 60\% of the total variance (see Fig.~\ref{fig:combined_heatmap_pca}). On the CIFAR-10 and CIFAR-100 datasets, we set the number of gaussian components to 10 and 100, respectively, and the PCA dimension to 60. For CleanSTL-10, which contains more than 10 categories in its training set, we set the number of gaussian components to 15 for improved robustness, the PCA dimension to 45, and the convergence threshold (Tol) to $10^1$. Classification performance is evaluated using accuracy. All experiments are conducted on a single NVIDIA Tesla T4 accelerator with 15~GB of memory.

\subsection{Comparison with State-of-the-art Methods}\label{subsection:comparison_with_state_of_the_art_methods}

\paragraph{Results on CIFAR-10.}
We show the benchmark results on CIFAR-10 in Table \ref{tab:cifar10_results}. Our method achieves competitive performance compared with state-of-the-art methods, including FlexMatch~\cite{zhang_flexmatch_2021}, SequenceMatch~\cite{Nguyen_2024_SequenceMatch} and EPASS~\cite{Nguyen_2024_epass}. It can be clearly seen that our method consistently outperforms other methods under all settings.

\begin{table}[H]
    \centering
    \caption{Benchmark results on CIFAR-10.}
    \label{tab:cifar10_results}
    \resizebox{0.8\textwidth}{!}
    {
    \begin{tabular}{lccc}
    \toprule
    Algorithms & Error Rate (40 labels) & Error Rate (250 labels) & Error Rate (4000 labels) \\
    \midrule
    Dash (\textcolor{green}{\citeyear[]{xu2021dash}}) & 9.29 \text{\footnotesize{$\pm$ 3.28}} & 5.16 \text{\footnotesize{$\pm$ 0.23}} & 4.36 \text{\footnotesize{$\pm$ 0.11}} \\
    MPL (\textcolor{green}{\citeyear[]{pham2021meta}}) & 6.62 \text{\footnotesize{$\pm$ 0.91}} & 5.76 \text{\footnotesize{$\pm$ 0.24}} & 4.36 \text{\footnotesize{$\pm$ 0.11}} \\
    FlexMatch (\textcolor{green}{\citeyear[]{zhang_flexmatch_2021}}) & 4.97 \text{\footnotesize{$\pm$ 0.06}} & 4.98 \text{\footnotesize{$\pm$ 0.09}} & 4.19 \text{\footnotesize{$\pm$ 0.01}} \\
    CoMatch (\textcolor{green}{\citeyear[]{li2021comatch}}) & 6.51 \text{\footnotesize{$\pm$ 1.18}} & 5.35 \text{\footnotesize{$\pm$ 0.14}} & 4.27 \text{\footnotesize{$\pm$ 0.12}} \\
    SimMatch (\textcolor{green}{\citeyear[]{zheng2022simmatch}}) & 5.38 \text{\footnotesize{$\pm$ 0.01}} & 5.36 \text{\footnotesize{$\pm$ 0.08}} & 4.41 \text{\footnotesize{$\pm$ 0.07}} \\
    AdaMatch (\textcolor{green}{\citeyear[]{roelofs2022adamatch}}) & 5.09 \text{\footnotesize{$\pm$ 0.21}} & 5.13 \text{\footnotesize{$\pm$ 0.05}} & 4.36 \text{\footnotesize{$\pm$ 0.05}} \\
    FreeMatch (\textcolor{green}{\citeyear[]{wang2023freematch}}) & 4.90 \text{\footnotesize{$\pm$ 0.12}} & 4.88 \text{\footnotesize{$\pm$ 0.09}} & 4.16 \text{\footnotesize{$\pm$ 0.06}} \\
    SoftMatch (\textcolor{green}{\citeyear[]{chen2023softmatch}}) & 5.11 \text{\footnotesize{$\pm$ 0.14}} & 4.96 \text{\footnotesize{$\pm$ 0.09}} & 4.27 \text{\footnotesize{$\pm$ 0.05}} \\
    SequenceMatch (\textcolor{green}{\citeyear[]{Nguyen_2024_SequenceMatch}}) & 4.80 \text{\footnotesize{$\pm$ 0.01}} & 4.75 \text{\footnotesize{$\pm$ 0.05}} & 4.15 \text{\footnotesize{$\pm$ 0.01}} \\
    EPASS (\textcolor{green}{\citeyear[]{Nguyen_2024_epass}}) & 5.55 \text{\footnotesize{$\pm$ 0.21}} & 5.31 \text{\footnotesize{$\pm$ 0.13}} & 4.23 \text{\footnotesize{$\pm$ 0.05}} \\
    \rowcolor{gray!20}\textbf{SemiOccam (Ours)} & \textbf{3.51 \text{\footnotesize{$\pm$ 0.12}}} & \textbf{3.47 \text{\footnotesize{$\pm$ 0.17}}} & \textbf{3.45 \text{\footnotesize{$\pm$ 0.16}}} \\
    \bottomrule
    \end{tabular}
    }
\end{table}

\paragraph{Results on CleanSTL-10.}
In Table \ref{tab:stl10_results}, we present the performance on CleanSTL-10. Our method achieves state-of-the-art performance, outperforming other methods by a large margin. Our method achieves the best performance in all three settings, demonstrating the effectiveness of our proposed method. Compared to SequenceMatch in \citeyear{Nguyen_2024_SequenceMatch}, our SemiOccam achieves +10.88\%, +8.35\%, and +1.41\% on 40-label, 250-label, and 1000-label settings, respectively.

\begin{table}[H]
    \centering
    \caption{Benchmark results on CleanSTL-10.}
    \label{tab:stl10_results}
    \resizebox{0.8\textwidth}{!}
    {
    \begin{tabular}{lccc}
        \toprule
        Algorithms & Error Rate (40 labels) & Error Rate (250 labels) & Error Rate (1000 labels)  \\
        \midrule
        Dash (\textcolor{green}{\citeyear[]{xu2021dash}}) & 42.00 \text{\footnotesize{$\pm$ 4.94}} & 10.50 \text{\footnotesize{$\pm$ 1.37}} & 6.30 \text{\footnotesize{$\pm$ 0.49}} \\
        MPL (\textcolor{green}{\citeyear[]{pham2021meta}}) & 35.97 \text{\footnotesize{$\pm$ 4.14}} & 9.90 \text{\footnotesize{$\pm$ 0.96}} & 6.66 \text{\footnotesize{$\pm$ 0.00}} \\
        FlexMatch (\textcolor{green}{\citeyear[]{zhang_flexmatch_2021}}) & 29.12 \text{\footnotesize{$\pm$ 5.04}} & 9.85 \text{\footnotesize{$\pm$ 1.35}} & 6.08 \text{\footnotesize{$\pm$ 0.34}} \\
        CoMatch (\textcolor{green}{\citeyear[]{li2021comatch}}) & 13.74 \text{\footnotesize{$\pm$ 4.20}} & 7.63 \text{\footnotesize{$\pm$ 0.94}} & 5.71 \text{\footnotesize{$\pm$ 0.08}} \\
        SimMatch (\textcolor{green}{\citeyear[]{zheng2022simmatch}}) & 16.98 \text{\footnotesize{$\pm$ 4.24}} & 8.27 \text{\footnotesize{$\pm$ 0.40}} & 5.74 \text{\footnotesize{$\pm$ 0.31}} \\
        AdaMatch (\textcolor{green}{\citeyear[]{roelofs2022adamatch}}) & 19.95 \text{\footnotesize{$\pm$ 5.17}} & 8.59 \text{\footnotesize{$\pm$ 0.43}} & 6.01 \text{\footnotesize{$\pm$ 0.02}} \\
        FreeMatch (\textcolor{green}{\citeyear[]{wang2023freematch}}) & 28.50 \text{\footnotesize{$\pm$ 5.41}} & 9.29 \text{\footnotesize{$\pm$ 1.24}} & 5.81 \text{\footnotesize{$\pm$ 0.32}} \\
        SoftMatch (\textcolor{green}{\citeyear[]{chen2023softmatch}}) & 22.23 \text{\footnotesize{$\pm$ 3.82}} & 9.18 \text{\footnotesize{$\pm$ 0.63}} & 5.79 \text{\footnotesize{$\pm$ 0.15}} \\
        SequenceMatch (\textcolor{green}{\citeyear[]{Nguyen_2024_SequenceMatch}}) & 15.45 \text{\footnotesize{$\pm$ 1.40}} & 12.78 \text{\footnotesize{$\pm$ 0.76}} & 5.56 \text{\footnotesize{$\pm$ 0.35}} \\
        EPASS (\textcolor{green}{\citeyear[]{Nguyen_2024_epass}}) & 9.15 \text{\footnotesize{$\pm$ 3.25}} & 6.27 \text{\footnotesize{$\pm$ 0.03}} & 5.40 \text{\footnotesize{$\pm$ 0.12}} \\
        \rowcolor{gray!20}\textbf{SemiOccam (Ours)} & \textbf{4.57 \text{\footnotesize{$\pm$ 0.24}}} & \textbf{4.43 \text{\footnotesize{$\pm$ 0.08}}} & \textbf{4.15 \text{\footnotesize{$\pm$ 0.07}}} \\
        \bottomrule
    \end{tabular}
    }
\end{table}

\paragraph{Results on CIFAR-100.}
In Table \ref{tab:cifar100_results}, we compare the performance of SemiOccam with state-of-the-art methods on CIFAR-100 dataset. Our method outperforms other methods in the 400 labels and 2,500 labels setting, and achieves comparable 10,000 labels settings.

\begin{table}[H]
    \centering
    \caption{Benchmark results on CIFAR-100.}
    \label{tab:cifar100_results}
    \resizebox{0.8\textwidth}{!}
    {
    \begin{tabular}{lccc}
        \toprule
        Algorithms & Error Rate (400 labels) & Error Rate (2500 labels) & Error Rate (10000 labels) \\
        \midrule
        Dash (\textcolor{green}{\citeyear[]{xu2021dash}}) & 44.82 \text{\footnotesize{$\pm$ 0.96}} & 27.15 \text{\footnotesize{$\pm$ 0.22}} & 21.88 \text{\footnotesize{$\pm$ 0.07}} \\
        MPL (\textcolor{green}{\citeyear[]{pham2021meta}}) & 46.26 \text{\footnotesize{$\pm$ 1.84}} & 27.71 \text{\footnotesize{$\pm$ 0.19}} & 21.74 \text{\footnotesize{$\pm$ 0.09}} \\
        FlexMatch (\textcolor{green}{\citeyear[]{zhang_flexmatch_2021}}) & 39.94 \text{\footnotesize{$\pm$ 1.62}} & 26.49 \text{\footnotesize{$\pm$ 0.20}} & 21.90 \text{\footnotesize{$\pm$ 0.15}} \\
        CoMatch (\textcolor{green}{\citeyear[]{li2021comatch}}) & 53.41 \text{\footnotesize{$\pm$ 2.36}} & 29.78 \text{\footnotesize{$\pm$ 0.11}} & 22.11 \text{\footnotesize{$\pm$ 0.22}} \\
        SimMatch (\textcolor{green}{\citeyear[]{zheng2022simmatch}}) & 39.32 \text{\footnotesize{$\pm$ 0.72}} & 26.21 \text{\footnotesize{$\pm$ 0.37}} & 21.50 \text{\footnotesize{$\pm$ 0.11}} \\
        AdaMatch (\textcolor{green}{\citeyear[]{roelofs2022adamatch}}) & 38.08 \text{\footnotesize{$\pm$ 1.35}} & 26.66 \text{\footnotesize{$\pm$ 0.33}} & 21.99 \text{\footnotesize{$\pm$ 0.15}} \\
        FreeMatch (\textcolor{green}{\citeyear[]{wang2023freematch}}) & 39.52 \text{\footnotesize{$\pm$ 0.01}} & 26.22 \text{\footnotesize{$\pm$ 0.08}} & 21.81 \text{\footnotesize{$\pm$ 0.17}} \\
        SoftMatch (\textcolor{green}{\citeyear[]{chen2023softmatch}}) & 37.60 \text{\footnotesize{$\pm$ 0.24}} & 26.39 \text{\footnotesize{$\pm$ 0.38}} & 21.86 \text{\footnotesize{$\pm$ 0.16}} \\
        SequenceMatch (\textcolor{green}{\citeyear[]{Nguyen_2024_SequenceMatch}}) & 37.86 \text{\footnotesize{$\pm$ 1.07}} & 25.99 \text{\footnotesize{$\pm$ 0.22}} & \textbf{20.10 \text{\footnotesize{$\pm$ 0.04}}} \\
        EPASS (\textcolor{green}{\citeyear[]{Nguyen_2024_epass}}) & 38.88 \text{\footnotesize{$\pm$ 0.24}} & 25.68 \text{\footnotesize{$\pm$ 0.33}} & 21.32 \text{\footnotesize{$\pm$ 0.14}} \\
        \rowcolor{gray!20}\textbf{SemiOccam (Ours)} & \textbf{26.59 \text{\footnotesize{$\pm$ 1.02}}} & \textbf{22.19 \text{\footnotesize{$\pm$ 0.81}}} & 21.21 \text{\footnotesize{$\pm$ 0.26}} \\
        \bottomrule
    \end{tabular}
    }
\end{table}

\subsection{Ablation Study}\label{subsection:ablation_study}

We conduct comprehensive ablation experiments to evaluate the effectiveness of different components in our SemiOccam framework. Table~\ref{tab:ablation} summarizes the classification error rates across three datasets under various settings.

\subsubsection{Effectiveness of SGMM vs. Softmax Classifier}\label{paragraph:comparing_semi_supervised_learning_methods}

To evaluate the benefit of using the SGMM component, we replace it with the standard ViT classification head based on a softmax layer and train both models under limited-label settings. The results in Table~\ref{tab:ablation} show that SemiOccam significantly outperforms the softmax-based head, especially in extremely low-label regimes.

\paragraph{ViT Classification Head for Semi-Supervised Learning.}\label{sec:vit_origin_classification_head}

We define the original ViT head as follows: Given the feature vector $\mathbf{f}_i \in \mathbb{R}^d$ extracted from the backbone, a linear projection is applied as $\mathbf{s}_i = \mathbf{W}_h \mathbf{f}_i + \mathbf{b}_h$, followed by a softmax function:
\begin{equation*}
P(c=k|\mathbf{f}_i) = \frac{\exp(s_{i,k})}{\sum_{j=1}^K \exp(s_{i,j})}, \quad \forall k \in \{1,\ldots,K\}
\end{equation*}
We adopt a confidence-based pseudo-labeling scheme to construct the unsupervised loss. The overall training objective combines both supervised and unsupervised loss terms:
\[
\mathcal{L}_{\text{total}} = \mathcal{L}_{\text{sup}} + \lambda(t) \cdot \mathcal{L}_{\text{unsup}}, \quad \lambda(t) = \lambda_{\text{max}} \cdot \min\left(1, \frac{t}{T_{\text{ramp}}}\right)
\]

\subsubsection{Impact of Pseudo-Labeling}\label{paragraph:effect_of_pseudo_labeling}

We further examine the effect of pseudo-labeling by disabling it in training. The row labeled “w/o P-L” in Table~\ref{tab:ablation} shows that removing pseudo-labeling leads to degraded performance, highlighting its importance for leveraging unlabeled data effectively.

\begin{table}[H]
    \centering
    \caption{Ablation results on CIFAR-10/100 and CleanSTL-10.}
    \label{tab:ablation}
    \resizebox{\textwidth}{!}{%
    \begin{tabular}{lccc|ccc|ccc}
    \toprule
    Dataset & \multicolumn{3}{c|}{CIFAR-10} & \multicolumn{3}{c|}{CIFAR-100} & \multicolumn{3}{c}{CleanSTL-10} \\
    \cmidrule(lr){2-4} \cmidrule(lr){5-7} \cmidrule(lr){8-10}
    \# Labels & 40 & 250 & 4000 & 400 & 2500 & 10000 & 40 & 250 & 1000 \\
    \midrule
    w Softmax   & 51.05 \text{\footnotesize{$\pm$ 16.69}} & 5.40 \text{\footnotesize{$\pm$ 1.70}} & 3.75 \text{\footnotesize{$\pm$ 0.21}} & 57.32 \text{\footnotesize{$\pm$ 10.91}} & 43.01 \text{\footnotesize{$\pm$ 5.89}} & 33.70 \text{\footnotesize{$\pm$ 2.71}} & 16.29 \text{\footnotesize{$\pm$ 2.85}} & 4.88 \text{\footnotesize{$\pm$ 1.50}} & 4.38 \text{\footnotesize{$\pm$ 0.40}} \\
    w/o P-L     & 3.73 \text{\footnotesize{$\pm$ 0.21}} & 3.70 \text{\footnotesize{$\pm$ 0.23}} & 3.63 \text{\footnotesize{$\pm$ 0.19}} & 30.56 \text{\footnotesize{$\pm$ 0.80}} & 23.21 \text{\footnotesize{$\pm$ 0.66}} & 21.56 \text{\footnotesize{$\pm$ 0.43}} & 4.91 \text{\footnotesize{$\pm$ 0.12}} & 4.81 \text{\footnotesize{$\pm$ 0.12}} & 4.58 \text{\footnotesize{$\pm$ 0.10}} \\
    \rowcolor{gray!20}SemiOccam   & \textbf{3.51} \text{\footnotesize{$\pm$ 0.12}} & \textbf{3.47} \text{\footnotesize{$\pm$ 0.17}} & \textbf{3.45} \text{\footnotesize{$\pm$ 0.16}} & \textbf{26.59} \text{\footnotesize{$\pm$ 1.02}} & \textbf{22.19} \text{\footnotesize{$\pm$ 0.81}} & \textbf{21.21} \text{\footnotesize{$\pm$ 0.26}} & \textbf{4.44} \text{\footnotesize{$\pm$ 0.04}} & \textbf{4.43} \text{\footnotesize{$\pm$ 0.08}} & \textbf{4.15} \text{\footnotesize{$\pm$ 0.07}} \\
    \bottomrule
    \end{tabular}%
    }
\end{table}

\subsection{Training Cost Analysis}\label{subsection:training_cost}

In the field of semi-supervised learning, besides model performance, training overhead is a key metric for evaluating the practicality and environmental impact of algorithms. This section will quantitatively compare the training overhead of existing benchmarks with our proposed method.

\textbf{TorchSSL} from \citeauthor{zhang_flexmatch_2021} is a mainstream SSL evaluation protocols, typically employ training deep neural networks from scratch, leading to significant computational burdens. For example, training the FixMatch algorithm on a single \textbf{NVIDIA V100 GPU} requires ``300 hours $\times$ 3 settings $\times$ 3 seeds'' for the CIFAR-100 dataset; ``110 hours $\times$ 3 settings $\times$ 3 seeds'' for CIFAR-10; and ``225 hours $\times$ 3 settings $\times$ 3 seeds'' for STL-10. The total training time amounts to \textbf{5715 GPU hours} (approximately 238 GPU days). This high computational demand poses a significant experimental cost barrier for academia and resource-constrained research institutions.

\textbf{USB} from \citeauthor{wang2022usb} is another current mainstream SSL evaluation protocol. It reports that training the FixMatch algorithm on a single \textbf{NVIDIA V100 GPU} incurs an overhead of ``11 hours $\times$ 2 settings $\times$ 3 seeds'' for the CIFAR-100 dataset, and ``18 hours $\times$ 2 settings $\times$ 3 seeds'' for the STL-10 dataset. Across these two datasets, a total of \textbf{174 GPU hours are required}.

\begin{figure}[ht]
    \centering
    \begin{minipage}[b]{0.8\textwidth}
        \centering
        \includegraphics[width=\textwidth]{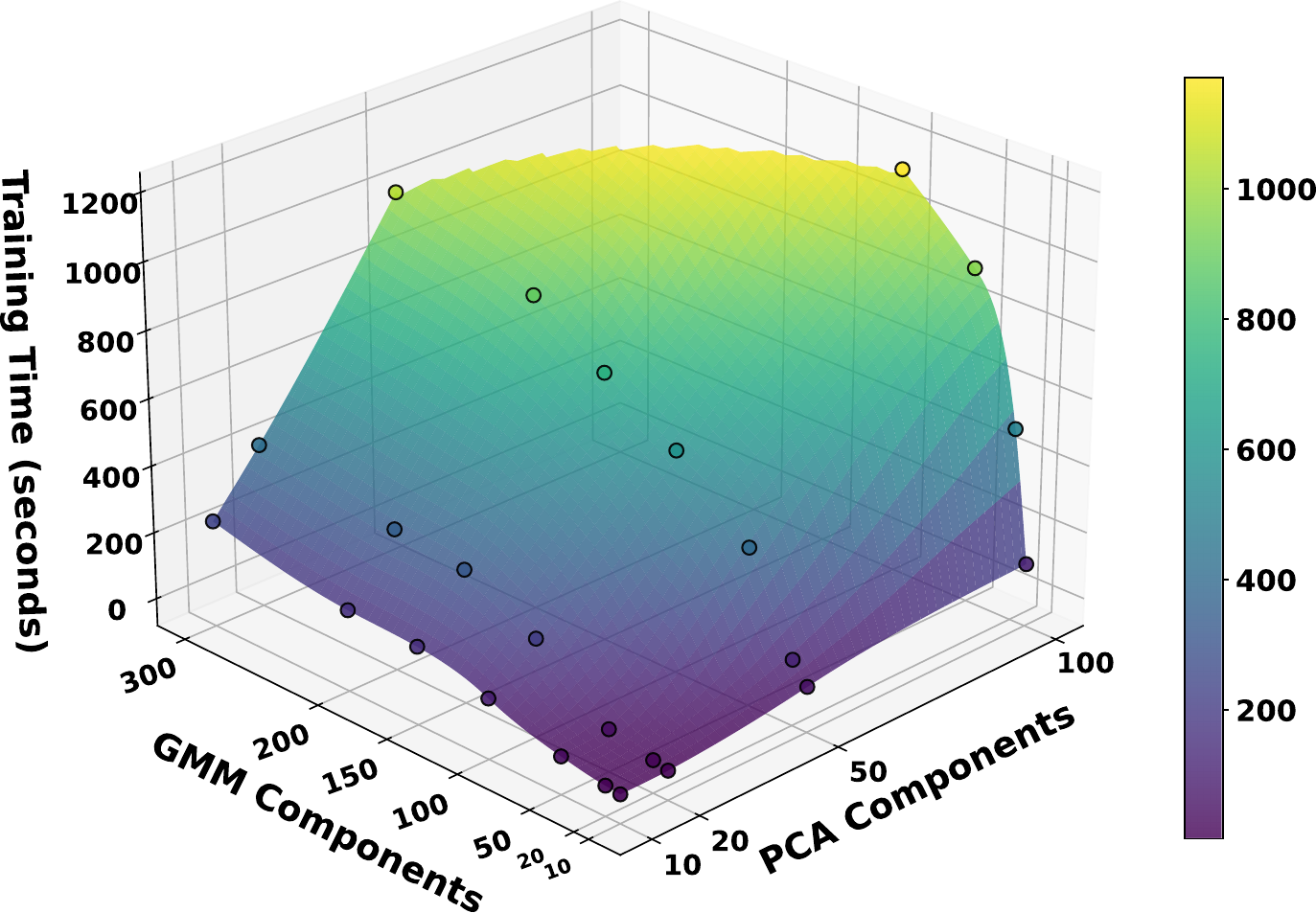}
    \end{minipage}
    \caption{Our Core Training Time Analysis. This surface plot illustrates the training time (in seconds) of the SGMM classifier under varying numbers of GMM and PCA components, assuming a 60,000-sample training dataset.}
    \label{fig:time}
\end{figure}

\textbf{Our proposed method} utilizes a pre-trained ViT as a feature extractor. After this feature extraction step is completed, our core classifier training process can typically be completed in \textbf{less than ten minutes} on a \textbf{single free Tesla T4 accelerator} on the Google Colab platform, demonstrating high efficiency and state-of-the-art performance. This minute-level training time gives our method a significant competitive advantage in \textbf{resource-constrained practical deployments} and \textbf{rapid iteration research scenarios}. For detailed training time performance, please refer to Figure \ref{fig:time}.

\section{Conclusions}\label{sec:conclusions}

We introduce high-performance feature extractors into the classical yet often underappreciated MoE classifier for image recognition tasks. Our architecture is simple and \textbf{training is highly efficient}, making it particularly \textbf{well-suited for scenarios with limited labeled data}. The proposed unified modeling strategy performs surprisingly well when trained with very few labeled samples and large-scale unlabeled data. In our study, we also \textbf{identify a long-overlooked data leakage issue} in the STL-10 dataset and \textbf{release the deduplicated CleanSTL-10 dataset} to ensure fair and reliable evaluation in semi-supervised learning. Extensive experiments show that SemiOccam \textbf{achieves new performance breakthroughs} on benchmarks such as CleanSTL-10 and CIFAR-10, maintaining over 96\% classification accuracy even with only 4 labeled samples per class \textbf{in under ten minutes} on a single Tesla T4 accelerator, compared to the hundreds of GPU hours required by existing methods.

\textbf{Future development} of this technology can proceed along several dimensions. \textbf{First}, constructing dynamic component adjustment mechanism based on a Bayesian framework to achieve parameter adaptivity. \textbf{Secondly}, enhancing the model's discriminative performance when facing complex tasks such as fine-grained classification. \textbf{Most interesting}, exploring its application in other fields, such as electroencephalogram signal analysis and medical image processing, thereby broadening its application scope and verifying its transferability. 

\textbf{We believe} these improvements will further unlock the technological benefits brought about by the fusion of probabilistic graphical models and deep learning. \textbf{Crucially}, given our method's advantages in short training times and high efficiency, we are particularly keen to see its widespread \textbf{adoption in edge deployment scenarios} and \textbf{applications requiring rapid iteration}.

\ifpaperfinal\makeacknowledgments{The work is supported by the National Natural Science Foundation of China No. 62076078, Fundamental Research Funds for the Central Universities No.3072024LJ0403, and the CAAI-Huawei MindSpore Open Fund No. CAAIXSJLJJ-2020-033A.}\fi

\bibliography{references}
\bibliographystyle{template}
\nocite{xing_multi-manifold_2013,chopin_performancegaussianmixturemodel_2024,porcher2024betterpseudolabelssemisupervisedinstance,mo2023sclipsemisupervisedvisionlanguagelearning,saberi2024outofdomainunlabeleddataimproves,takahashi2024rolepseudolabelsselftraininglinear,majurski2024methodmomentsembeddingconstraint,jing2022understanding}

\end{document}